# FPGA Architecture for Deep Learning and its application to Planetary Robotics

Pranay Reddy Gankidi
Space and Terrestrial Robotic
Exploration (SpaceTREx) Lab
Arizona State University
781 E. Terrace Rd., AZ 85287
pgankidi@asu.edu

Jekan Thangavelautham
Space and Terrestrial Robotic
Exploration (SpaceTREx) Lab
Arizona State University
781 E. Terrace Rd., AZ 85287
jekan@asu.edu

*Abstract*— Autonomous control systems onboard planetary rovers and spacecraft benefit from having cognitive capabilities like learning so that they can adapt to unexpected situations in-situ. Q-learning is a form of reinforcement learning and it has been efficient in solving certain class of learning problems. However, embedded systems onboard planetary rovers and spacecraft rarely implement learning algorithms due to the constraints faced in the field, like processing power, chip size, convergence rate and costs due to the need for radiation hardening. These challenges present a compelling need for a portable, low-power, area efficient hardware accelerator to make learning algorithms practical onboard space hardware. This paper presents a FPGA implementation of Q-learning with Artificial Neural Networks (ANN). This method matches the massive parallelism inherent in neural network software with the fine-grain parallelism of an FPGA hardware thereby dramatically reducing processing time. Mars Science Laboratory currently uses Xilinx-Space-grade Virtex FPGA devices for image processing, pyrotechnic operation control and obstacle avoidance. We simulate and program our architecture on a Xilinx Virtex 7 FPGA. The architectural implementation for a single neuron Q-learning and a more complex Multilayer Perception (MLP) Q-learning accelerator has been demonstrated. The results show up to a 43-fold speed up by Virtex 7 FPGAs compared to a conventional Intel i5 2.3 GHz CPU. Finally, we simulate the proposed architecture using the Symphony simulator and compiler from Xilinx, and evaluate the performance and power consumption.

## Table of Contents

## 1. Introduction

Space missions can benefit from machine learning algorithms to increase operational autonomy. Missions on Mars and beyond are faced with long latencies that prevent teleoperation. Typically, machine learning algorithms are computationally intensive and require significant processing power, making it a concern for running such algorithms on planetary rovers and space hardware [2]. However, autonomous software is being used. Examples of autonomous software present on Mars Science Laboratory (MSL) rover is AEGIS (Autonomous Exploration for Gathering Increased Science) software, which helps the rover to autonomously choose targets for the laser [3] as shown in Figure 1. Other forms of autonomous software are intended for use in vision, particularly particle detection [26-27] and for localization.

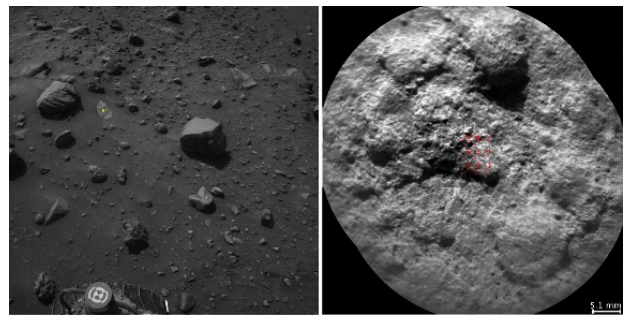

Figure 1. Target selection and pointing based on Navcam Imagery on MSL Rover [3].

Reinforcement Learning (RL) [4] is one potential learning algorithm and it determines how an agent in a random un-modelled environment ought to take actions to maximize the cumulative future reward. Through trial-and-error interactions with its environment, RL aids a robot to autonomously discovering an optimal behavior. In RL, human interactions with the robot is confined to provide the feedback in terms of an objective function that measures the one-step performance of the robot rather than obtaining a generalized closed-form solution [5]. This is one of the key advantages of RL because in real-world applications there is seldom any chance of finding a closed form solution to a learning problem. The goal in RL is specified by the reward function, which either acts as reinforcement or punishment that is dependent on the performance of the autonomous agent with respect to the desired goal.

Reinforcement Learning has been in existence for 30 years, however the most recent innovations of combining learning with deep neural networks [6] is shown to be human competitive. This new approach has been proven to beat humans in various video games [7]. One of the implemented examples is shown in Figure 2. Some of the other examples in robotic applications include, robots learning how to perform complicated manipulation tasks, like the terrain adaptive locomotion skills using deep reinforcement learning [8] and Berkley robot performing new tasks like a child [9]. Robot platforms running such algorithms have large computational demands and often have long execution times when implemented using traditional CPUs [10]. This challenge has paved the way for development of accelerators with many parallel cores that can in theory speed up computation for low-power. These devices consist of GPUs, ASICs and FPGAs.

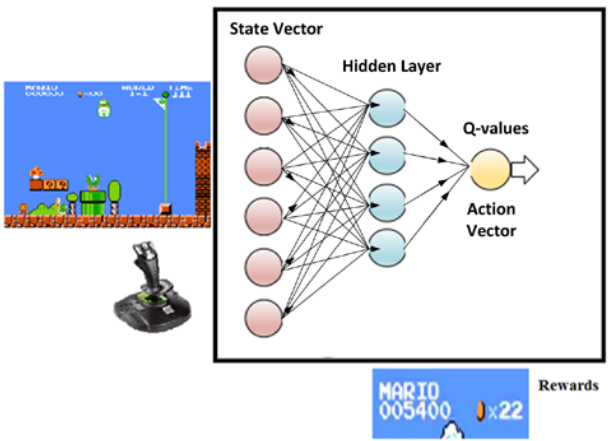

**Figure 2. Playing Mario using Reinforcement learning.**

Over the past few years, we see a prevalence in cloud based deep learning which implements the deep learning algorithms in a cluster of high-processing computing units. Recently, Google developed an artificial brain which learns to find cat videos using a neural network that is built on 16,000 computer processors with more than one billion connections [11]. Such large servers often consume a lot of power. For example, one large, 50,000 square feet data center consumes around 5 MW of power (enough to power 5,000 houses) [12]. Power and radiation is a major concern for planetary rovers. MSL's onboard computers consist of RAD750, a radiation-hardened processor with a PowerPC architecture running at 133 MHz and having a throughput of 240 MIPS. This processor consumes 5 W of power. In comparison, the Nvidia Tegra K1 mobile processors specifically designed for low power computing with 192 Nvidia Cuda Cores consumes 12 W of power.

FPGAs are starting to emerge as a compelling alternative for GPUs in implementing neural network algorithms [13][14][15]. Microsoft's research paper on accelerating deep convolution neural networks using specialized hardware demonstrates a 2x performance improvement in accelerating Bing rankings [16]. In our work here, we combine deep neural networks with Q-learning and implement it on board an FPGA.

Alternate approaches include Neuro-evolution using algorithms like the Artificial Neural Tissue (ANT) [20-22] that have been used to control single and multi-robot systems for off-world applications [23-25]. ANT has been found to beat humans in the control of robot teams for excavation tasks [24]. ANT utilizes a neural-network controller much like the Q-learning approach presented here but is trained using Evolutionary Algorithms [20-22]. The Q-learning training approach utilizes back-propagation which requires calculating gradients and hence the error function needs to be differentiable. ANT however can work with non-differentiable, discontinous and discrete cases [21-22]. As with Q-learning, the process benefits from trial and error learning utilizing a goal function. The work presented here is directly applicable to neuro-evolution techniques including ANT as FPGA implementation of Q-learning using neural networks can be easily transferred and used for neuro-evolution.

The rest of the paper is organized as follows: In Section 2, we present the Q-learning algorithm (a subclass of Reinforcement Learning). In Section 3, we show how neural networks aid in improving the performance of Q-learning algorithm. In Section 4, we present the implementation architecture, and start with a single perceptron based architecture for Q-learning and continue with multilayer perceptron based Q-learning approach. Finally, in Section 5, we present the simulation results for the implemented architecture followed by conclusions.

## 2. Q-LEARNING ALGORITHM

Q-learning is a reinforcement learning technique that looks for and selects an optimal action based on an action selection function called the Q-function [1]. The Q function determines the utility of selecting an action, which takes into account optimal action (*a*) selected in a state (*s*) that leads to a maximum discounted future reward when optimal action is selected from that point. The Q-function can be represented as follows:

$$Q(s_t, a_t) = \max R_{t+1} \quad (1)$$

The future rewards are obtained through continuous iterations selecting one of the actions based on existing Q-values, performing an action in the future and updating the Q-function in the current state. Let us assume that we pick an optimal action in the future based on the maximum Q-value. The equation demonstrates the action policy selected based on the maximum Q-value:

$$\pi(s) = \arg(max\, R_{t+1}) \quad (2)$$

After selecting an optimal action, we move on to a new state, obtaining rewards during this process. Let us assume that new state to be $s_{t+1}$, The optimal value for next state is found out by iterating through all the Q-values in next state and finding an action $a_{t+1}$ which produces an optimal Q-value in a state.

$$opt\ Q(t+1) = max_{a'}\ Q(s_{t+1}, a'_{t+1}) \quad (3)$$

Now that we have the optimal Q-value for next state, we update the current Q-value with the following famous Q-update function

$$Q(s,a) = Q(s,a) + \propto [r + \gamma . opt\ Q(t+1) - Q(s,a)] \quad (4)$$

The above equation is the basis of the Q-learning algorithm, in which $\propto$ is the learning rate or learning factor set between 0 and 1, that determines how much a Q-error update influences the current Q-values. In general, when learning rate is large we might overshoot the local minimum of the Q-function. At a learning rate equal to 0, the value of Q never updates and remains same as before thereby stopping learning. At a learning rate equal to 1, we see the Q-values cancelling out updates of the Q-value with its target value

Q-learning with neural networks eliminates the usage of the Q-table as the neural network acts as a Q-function solver [17]. Instead of storing all the possible Q-values, we estimate the Q-value based on the output of the neural network, which has been trained with Q-value errors. Q-values for a given state-action pair can be stored using a single neural network. To find the Q-value for a given state and action, we perform a feed-forward calculation for all actions possible in a state, thus calculating the Q-values for each of the actions. Based on the calculated Q-value, an action is chosen using one of the action selection policies. This action selection is performed in real-time, which leads to a new state. The feed-forward step is run again, this time with next state as its input and for all possible actions in the next state. The target Q-value is calculated using Equation 4, which is used to propagate the error in the neural network.

The state-flow for the Q-learning algorithm using neural networks to update a single Q-value for a state-action pair is:
(1) Calculate Q-values for all actions in a state by running the feed-forward step 'A' times where A is the number of possible actions in a state.
(2) Select an action ($a_t$), and move to a next state ($s_{t+1}$).
(3) Calculate Q-values for all actions in next state ($s_{t+1}$) by running the feed-forward step 'A' times where A is the number of possible actions in next state.
(4) Determine the error value as in Equation 4, by calculating the maximum of $q$ values in next state.
(5) Use the error signal to run a backpropagation step, to update weights and biases.

## 3. PERCEPTRON Q LEARNING ACCELERATOR

A Perceptron is as a single neuron in a multi-layer neural network. It has a multi-dimensional input and a single output. A perceptron contains weights for each of the inputs and a single bias, as shown in Figure 3.

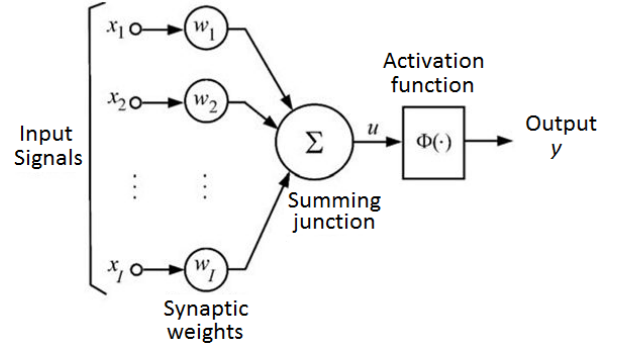

**Figure 3. Perceptron architecture**

The weighted sum of all the inputs is calculated using the Equation 5, which is modelled as a combination of a multiplier and an accumulator in hardware:

$$\sigma = (\Sigma_{i \in N}\ x_i w_i) \quad (5)$$

The output of a perceptron, also called the firing rate is calculated by passing the value of σ through the activation function as follows:

$$O = f(\sigma) = \frac{1}{1+e^{-\sigma}} \quad (6)$$

There are many number of hardware schematics existing for implementing the activation function [18-19]. We utilize a Look-up Table approach, which stores the pre-calculated values of the sigmoid values. The size of ROM plays a major role in the accuracy of the output value. As the sensitivity of the stored values increases, the lookup time increase.

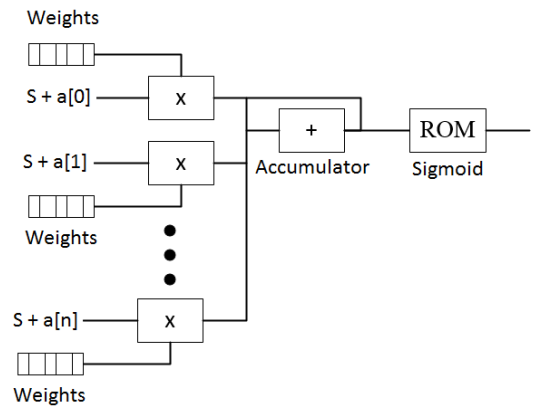

**Figure 4. Feed Forward Step Hardware Schematic**

The feed forwards step is run twice for updating one single Q-value, once to calculate the q values in current state, and then to calculate the q values in next state. During step 4, error is determined by buffering out all the FIFO Q-values of the current and next state in parallel by calculating the maximum value of next state Q-values, followed by applying Equation 4. The hardware schematic of our implementation is shown in Figure 5.

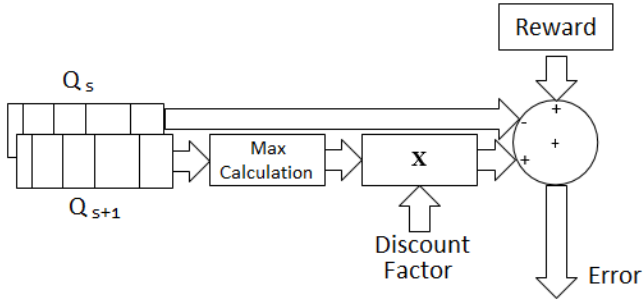

**Figure 5. Error Capture Hardware Schematic Block**

The control path and data implementation path for the module is shown in Figure 6. Two buffers have been implemented to store the Q-values for all actions in a state. One of them stores the Q-values for current state and the other stores Q-values for the next state. For a perceptron, a single backpropagation block is implemented to update the weights and biases of the neural network.

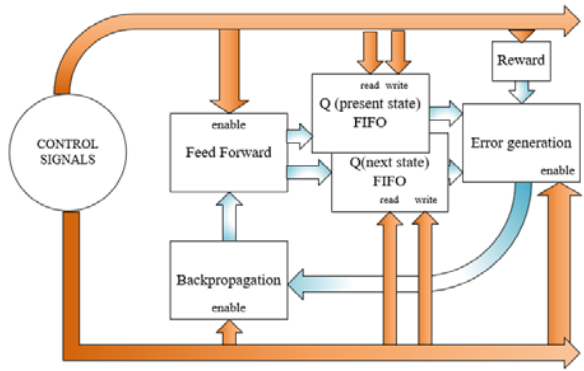

**Figure 6. Control and Data path for single neuron architecture**

Q-learning differs from other supervised learning from the fact that the backpropagation happens not through predefined errors but estimates of the error. The error value propagated is given by the Equation 7, where $f'(\sigma)$ is the derivative of a sigmoid function. The derivative of the sigmoid is also implemented using a Look-up Table (ROM) with pre-calculated set of values.

$$\delta = f'(\sigma)(Q_{error}) \quad (7)$$

Q-error is propagated backwards, and is used to train the neural network, the value of the Q-error is determined by Equation 8, where $Q(t+1)$ are the Q-values present in the next state buffer.

$$Q_{error} = \propto [r + \gamma . opt\ Q(t+1) - Q(s,a)] \quad (8)$$

The weights are updated using the following equation, where C is the learning factor

$$\Delta W = (C O \delta) \quad (9)$$

$$W_{new} = W_{prev} + \Delta W \quad (10)$$

The weights are updated by reading the weight values from the buffer, updating them using Equation 9 and writing them back to the FIFO buffer. A schematic of the implemented architecture for Q-learning is presented in Figure 7.

Throughput values differ between fixed and floating point implementation of the architecture as shown in Table 1. In a fixed point architecture, total number of clock cycles to update a single Q value equals $7A + 1$, with $A$ being the number of actions possible in a state. For an action size equal to 9, the total number of Q-values computed per second equals 2.34 million for a simple environment, and 0.53 Million for a complex environment.

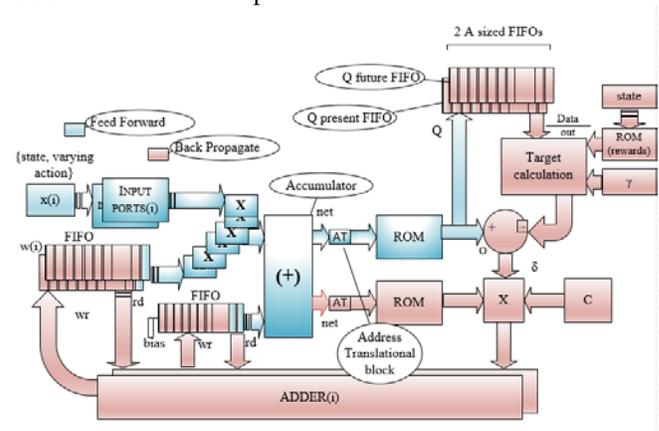

**Figure 7. Single Neuron Q-learning accelerator architecture**

**Table 1. Throughput calculation**

| Architecture | Throughput |
|---|---|
| Fixed Point Simple | 2340 kQ/second |
| Floating Point Simple | 290 kQ/second |
| Fixed Point Complex | 530 kQ/second |
| Floating Point Complex | 10 kQ/second |

## 4. MULTILAYER PERCEPTRON (MLP) Q LEARNING ACCELERATOR

The Q-learning architecture for a Multilayer Perceptron (MLP) is an extension of a single perceptron. The control path and data path is shown in Figure 8.

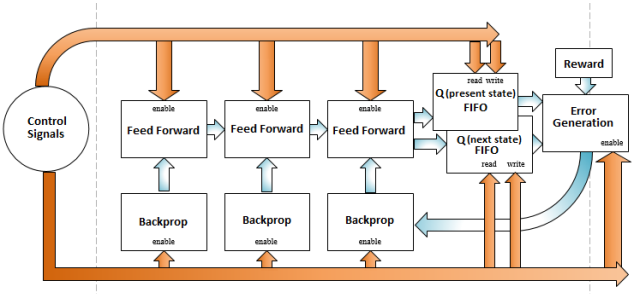

**Figure 8. Control and Data path for Q learning MLP accelerator architecture**

Each feedforward step is similar to that of Figure 4. Two buffers have been used with a size equal to the number of actions per state, to store the Q-values of current state and next state respectively. The feed forward block diagram for the complete MLP is shown in Figure 9. Though the error generation is similar in comparison to a single perceptron, the backpropagation block is different from the backpropagation of a single perceptron. The error value to be propagated is calculated using following equation.

$$\delta_i = f'(\sigma_i)(Q_{error}) \ \forall \ i, \ i \in outputLayer \qquad (11)$$

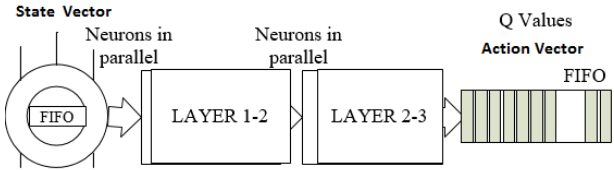

**Figure 9. Feed Forward Step for MLP Q learning**

Based on the output layer error value in Equation 11, error values for each of the hidden layers and input layer is calculated using Equation 12.

$$\delta_i = f'(\sigma_i) \ (\sum_{j \in outputLayer} \delta_j W_{ij})$$
$$\forall \ i \in outputLayer \qquad (12)$$

The weights of each hidden, input and output layer are updated using Equations 13 and 14, where $C$ is the learning rate of the neural network, $A$ and $B$ are the neurons in the current and next layer respectively.

$$\Delta W_{ij} = (CO_i \ \delta_j) \ \forall i, j \ ; \ i \in A \ ; j \in B \qquad (13)$$

$$W_{ij} = W_{ij(prev)} + \Delta W_{ij} \qquad (14)$$

Blocks for generating $\delta$ and $\Delta W$ are done using separate resources, thereby exploiting the fine-grained parallelism of the architecture. Figure 10 shows the backpropagation algorithm for Q-learning with weight updates happening using $\delta$ and $\Delta W$ generator. As seen earlier, for the MLP case (Table 2), throughput can vary significantly based on fixed vs. floating point precision and neuron complexity.

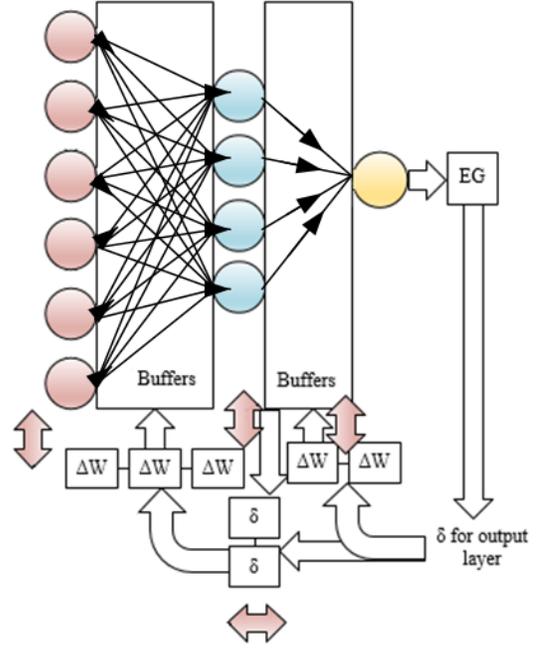

**Figure 10. Backpropagation for Q-learning MLP**

**Table 2. Throughput calculation for MLP**

| Architecture | Throughput |
| --- | --- |
| Fixed Point Simple | 1060 kQ/second |
| Floating Point Simple | 745 kQ/second |
| Fixed Point Complex | 247 kQ/second |
| Floating Point Complex | 9 kQ/second |

## 5. RESULTS

The presented architectures are simulated using Xilinx Tools on Vertex 7 FPGA. The following combinations of architecture and environment have been considered.

(1) Single neuron in a simple environment,
(2) Single neuron in a complex environment,
(3) Multilayer perceptron in a simple environment,
(4) Multilayer perceptron in a complex environment

Simple and complex environment vary by the fact that the simple environment has a small size of state, action vector and number of possible actions per state. In our case the size is equal to 6 with the size of the state vector equal to 4 and size of the action vector equal to 2. The complex environment is modelled with the total size of the state and action vector as 20, possible number of actions per state as 40, and the state space size as 1800. The neural network architecture for MLP consists of 11 neurons in a simple environment and 25 neurons in a complex environment with 4 hidden layer neurons.

Table 3 through 6 presents the time taken to update 1 single Q-value for each of the architecture implementations. The Fixed point parallel architectures have high performance. We note up to a 95-fold improvement in simulating a single neuron going from a conventional Intel i5 6$^{th}$ Gen 2.3 GHz CPU to a Xilinx Virtex-7 FPGA. Further, if we scale up to a multilayer perceptron, Virtex-7 FPGA utilizing fixed-point achieves up to a 43-fold improvement over a conventional CPU. The major advantage comes from utilizing fixed point representation. Improvements are expected even with floating point use but are not significant. These results show it is critical to transform the problem into a fixed point representation to exploit the advantage of a Virtex-7 FPGA.

The fixed point word length and fraction length plays a major role in trading off accuracy with power consumption. Based on this fact, fixed-point architecture can be implemented with high accuracy, same throughput or performance metric as that of Figure 2, while having increased power consumption.

**Table 3: Simple Neuron**

| Architecture | Completion Time (µs) | Advantage |
|---|---|---|
| FPGA – Virtex 7, Fixed | 0.4 | 22x |
| FPGA – Virtex 7, Floating | 7.7 | 1.5x |
| CPU – Intel i5 2.3 GHz | 20 | 1x |

**Table 4: Complex Neuron**

| Architecture | Completion Time (µs) | Advantage |
|---|---|---|
| FPGA – Virtex 7, Fixed | 1.8 | 95x |
| FPGA – Virtex 7, Floating | 102 | 1.7x |
| CPU – Intel i5 2.3 GHz | 172 | 1x |

**Table 5: Simple Multilayer Perceptron (MLP)**

| Architecture | Completion Time (µs) | Advantage |
|---|---|---|
| FPGA – Virtex 7, Fixed | 0.9 | 22x |
| FPGA – Virtex 7, Floating | 13 | 1.5x |
| CPU – Intel i5 2.3 GHz | 20 | 1x |

**Table 6: Complex Multilayer Perceptron (MLP)**

| Architecture | Completion Time (µs) | Advantage |
|---|---|---|
| FPGA – Virtex 7, Fixed | 4 | 43x |
| FPGA – Virtex 7, Floating | 107 | 1.6x |
| CPU – Intel i5 2.3 GHz | 172 | 1x |

We observe in Table 7 and 8, that the peak power consumption is slightly higher for floating point architecture when simulating a multilayer perceptron at 150 MHz frequency. Though power estimation is an important factor for consideration, the energy values is what that is most useful for comparisons, since Q-values change over time with improving accuracy, the total time taken to calculate the optimal Q values vary for each of the architectures. The total time taken for finding the optimal Q values can only be obtained when the learning algorithm is implemented on real space hardware and is exposed to simple and complex environments. Overall, FPGAs point towards a promising pathway to implement learning algorithms for use on aerospace robotic platforms.

**Table 7: Power Consumption for Simple Multilayer Perceptron (MLP)**

| Architecture | Power (w) | Advantage |
|---|---|---|
| FPGA – Virtex 7, Fixed | 5.6 | 1.3x |
| FPGA – Virtex 7, Floating | 7.1 | 1x |

**Table 8: Power Consumption for Complex Multilayer Perceptron (MLP)**

| Architecture | Power (w) | Advantage |
|---|---|---|
| FPGA – Virtex 7, Fixed | 7.1 | 1.3x |
| FPGA – Virtex 7, Floating | 10 | 1x |

## 6. CONCLUSION

Fine grained parallelism for Q-learning using single perceptron and multilayer perceptron has been implemented on a Xilinx Vertex-7 v485T FPGA. The results show up to a 43-fold speed up by Virtex 7 FPGAs compared to a conventional Intel i5 2.3 GHz CPUs. Fine grained parallel architectures are competitive in terms of area and power consumption and we observe a power consumption of 9.7W, for a multi-layer perceptron floating point FPGA. The power consumption can be further reduced by introducing pipelining in the data path. These results show the promise of parallelism in FPGAs. Our architecture matches the massive parallelism inherent in neural network software with the fine-grain parallelism of an FPGA hardware thereby dramatically reducing processing time. Further work will be done to apply this technology on single and multi-robot platforms in the laboratory followed by field demonstrations.

## BIOGRAPHY

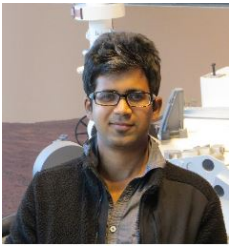

*Pranay Reddy Gankidi* received a B.E. in Electrical and Electronics Engineering from BITS PILANI, Pilani, India in 2013. He is presently pursuing his Masters in Computer Engineering from Arizona State University, AZ. He has interned at AMD with a focus on evaluating power efficient GPU architectures. He specializes in VLSI and Architecture, and his interests include design and evaluation of power and performance efficient hardware architectures and space exploration.

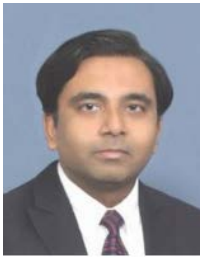

*Jekan Thangavelautham* is an Assistant Professor and has a background in aerospace engineering from the University of Toronto. He worked on Canadarm, Canadarm 2 and the DARPA Orbital Express missions at MDA Space Missions. Jekan obtained his Ph.D. in space robotics at the University of Toronto Institute for Aerospace Studies (UTIAS) and did his postdoctoral training at MIT's Field and Space Robotics Laboratory (FSRL). Jekan Thanga heads the Space and Terrestrial Robotic Exploration (SpaceTREx) Laboratory at Arizona State University. He is the Engineering Principal Investigator on the AOSAT I CubeSat Centrifuge mission and is a Co-Investigator on SWIMSat, an Airforce CubeSat mission to monitor space threats.